\documentclass{article}

\usepackage{times}
\usepackage{graphicx} 
\usepackage{subfigure} 

\usepackage{algorithm}
\usepackage{algorithmic}

\usepackage{hyperref}

\usepackage[accepted]{icml2017} 

\usepackage{dirtytalk}
\usepackage{pdfpages}
\usepackage{float}
\usepackage{amsmath}
\usepackage{amssymb}

\icmltitlerunning{Unimodal Probability Distributions for Deep Ordinal Classification}

\begin{document} 

\twocolumn[
\icmltitle{Unimodal Probability Distributions for Deep Ordinal Classification}



\icmlsetsymbol{equal}{*}

\begin{icmlauthorlist}
\icmlauthor{Christopher Beckham}{mila}
\icmlauthor{Christopher Pal}{mila}
\end{icmlauthorlist}

\icmlaffiliation{mila}{Montr\'eal Institute of Learning Algorithms, Qu\'{e}bec, Canada}

\icmlcorrespondingauthor{Christopher Beckham}{christopher.beckham@polymtl.ca}

\icmlkeywords{ordinal, unimodal, kappa, neural networks, deep learning, machine learning, ICML}

\vskip 0.3in
]



\printAffiliationsAndNotice{}  

\begin{abstract} 
Probability distributions produced by the cross-entropy loss for ordinal classification problems can possess undesired properties. We propose a straightforward technique to constrain discrete ordinal probability distributions to be unimodal via the use of the Poisson and binomial probability distributions. We evaluate this approach in the context of deep learning on two large ordinal image datasets, obtaining promising results.
\end{abstract} 

\section{Introduction}

Ordinal classification (sometimes called ordinal regression) is a prediction task in which the classes to be predicted are discrete and ordered in some fashion. This is different from discrete classification in which the classes are not ordered, and different from regression in that we typically do not know the distances between the classes (unlike regression, in which we know the distances because the predictions lie on the real number line). Some examples of ordinal classification tasks include predicting the stages of disease for a cancer \cite{gentry2015penalized}, predicting what star rating a user gave to a movie \cite{koren2011ordrec}, or predicting the age of a person \cite{eidinger2014age}.

\begin{figure*}
    \centering
    \subfigure[An adult woman]{\label{fig:abc_a}\includegraphics[width=0.15\textwidth]{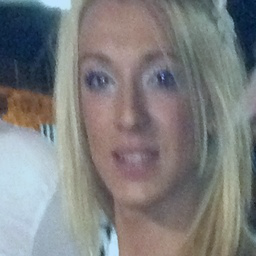}} \ \ \ \ \ \ \ \ \ \subfigure[Three probability distributions exhibiting the same mass for the `adult' class and therefore the same cross-entropy error.]{\label{fig:abc_b}\includegraphics[width=0.65\textwidth]{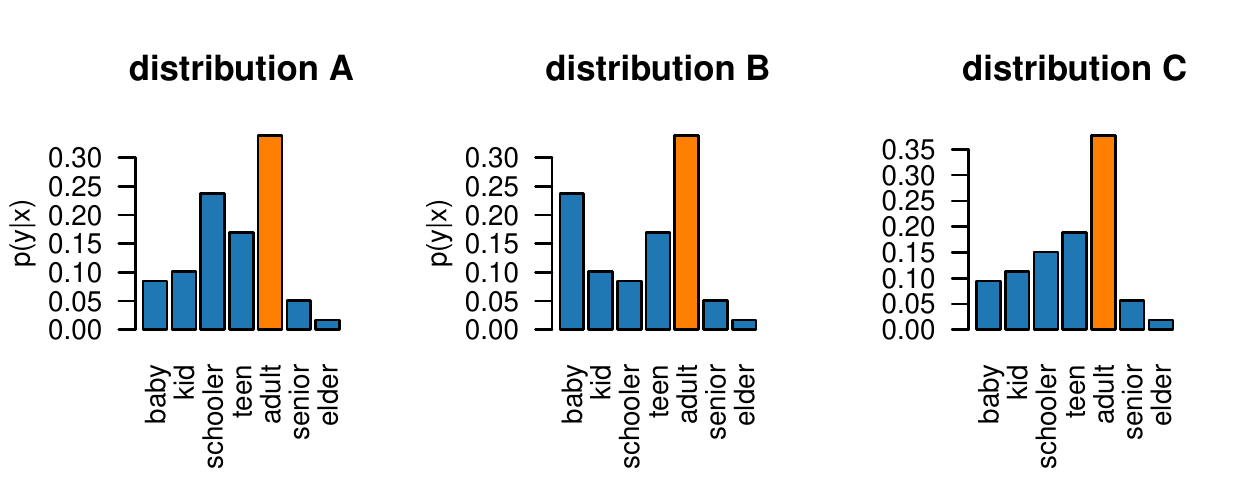}}
    \caption{Three ordinal probability distributions conditioned on an image of an adult woman. Distributions \textit{A} and \textit{B} are unusual in the sense that they are multi-modal.}
    \label{fig:abc}
\end{figure*}

Two of the easiest techniques used to deal with ordinal problems include either treating the problem as a discrete classification and minimising the cross-entropy loss, or treating the problem as a regression and using the squared error loss. The former ignores the inherent ordering between the classes, while the latter takes into account the distances between them (due to the square in the error term) but assumes that the labels are actually real-valued -- that is, adjacent classes are equally distant. Furthermore, the cross-entropy loss -- under a one-hot target encoding -- is formulated such that it only `cares' about the ground truth class, and that probability estimates corresponding to the other classes may not necessarily make sense in context. We present an example of this in Figure \ref{fig:abc}, showing three probability distributions: \textit{A}, \textit{B}, and \textit{C}, all conditioned on some input image. Highlighted in orange is the ground truth (i.e. the image is of an adult), and all probability distributions have identical cross-entropy: this is because the loss only takes into account the ground truth class, $-\log(p(y|\mathbf{x})_{c})$, where $c = adult$, and all three distributions have the same probability mass for the adult class.

Despite all distributions having the same cross-entropy loss, some distributions are `better' than others. For example, between \textit{A} and \textit{B}, \textit{A} is preferred, since \textit{B} puts an unusually high mass on the baby class. However, \textit{A} and \textit{B} are both unusual, because the probability mass does not gradually decrease to the left and right of the ground truth. In other words, it seems unusual to place more confidence on `schooler' than `teen' (distribution \textit{A}) considering that a teenager looks more like an adult than a schooler, and it seems unusual to place more confidence on 'baby' than 'teen' considering that again, a teenager looks more like an adult than a baby. Distribution \textit{C} makes the most sense because the probability mass gradually decreases as we move further away from the most confident class. In this paper, we propose a simple method to enforce this constraint, utilising the probability mass function of either the Poisson or binomial distribution.

For the remainder of this paper, we will refer to distributions like \textit{C} as `unimodal' distributions; that is, distributions where the probability mass gradually decreases on both sides of the class that has the majority of the mass.

\subsection{Related work}

Our work is inspired by the recent work of \citet{emd}, who shed light on the issues associated with different probability distributions having the same cross-entropy loss for ordinal problems. In their work, they alleviate this issue by minimising the `Earth mover's distance', which is defined as the minimum cost needed to transform one probability distribution to another. Because this metric takes into account the distances between classes -- moving probability mass to a far-away class incurs a large cost -- the metric is appropriate to minimise for an ordinal problem. It turns out that in the case of an ordinal problem, the Earth mover's distance reduces down to Mallow's distance:
\begin{equation} \label{eq:emd2}
    \text{emd}(\mathbf{\hat y}, \mathbf{y}) = \Big(\frac{1}{K}\Big)^{\frac{1}{l}} || \text{cmf}(\mathbf{\hat y}) - \text{cmf}(\mathbf{y}) ||_{l},
\end{equation}
where $\text{cmf}(\cdot)$ denotes the cumulative mass function for some probability distribution, $\mathbf{y}$ denotes the ground truth (one-hot encoded), $\mathbf{\hat y}$ the corresponding predicted probability distribution, and $K$ the number of classes.
The authors evaluate the EMD loss on two age estimation and one aesthetic estimation dataset and obtain state-of-the-art results. However, the authors do not show comparisons between the probability distributions learned between EMD and cross-entropy.

Unimodality has been explored for ordinal neural networks in \citet{da2008unimodal}. They explored the use of the binomial and Poisson distributions and a non-parametric way of enforcing unimodal probability distributions (which we do not explore). One key difference between their work and ours here is that we evaluate these unimodal distributions in the context of deep learning, where the datasets are generally much larger and have more variability; however, there are numerous other differences which we will highlight throughout this paper.

\citet{mypaper} explored a loss function with an intermediate form between a cross-entropy and regression loss. In their work the squared error loss is still used, but a probability distribution over classes is still learned. This is done by adding a regression layer (i.e. a one-unit layer) at the top of what would normally be the classification layer, $p(y|\mathbf{x})$. Instead of learning the weight vector $\mathbf{a}$ it is fixed to $[0, \dots, K-1]^{T}$ and the squared error loss is minimised. This can be interpreted as drawing the class label from a Gaussian distribution $p(c|\mathbf{x}) = N(c; \mathbb{E}[\mathbf{a}]_{p(y|\mathbf{x})}, \sigma^{2})$. This technique was evaluated against the diabetic retinopathy dataset and beat most of the baselines employed. Interestingly, since $p(c|\mathbf{x})$ is a Gaussian, this is also unimodal, though it is a somewhat odd formulation as it assumes $c$ is continuous when it is really discrete.

\citet{cheng} proposed the use of binary cross-entropy or squared error on an ordinal encoding scheme rather than the one-hot encoding which is commonly used in discrete classification. For example, if we have $K$ classes, then we have labels of length $K-1$, where the first class is $[0, \dots, 0]$, second class is $[1, \dots, 0]$, third class is $[1, 1, \dots, 0]$ and so forth. With this formulation, we can think of the $i$'th output unit as computing the cumulative probability $p(y > i | \mathbf{x})$, where $i \in \{0, \dots, K-2\}$. \citet{frank} also proposed this scheme but in a more general sense by using multiple classifiers (not just neural networks) to model each cumulative probability, and \citet{niu2016ordinal} proposed a similar scheme using CNNs for age estimation. This technique however suffers from the issue that the cumulative probabilities $p(y > 0  \ | \ \mathbf{x}), \dots, p(y > K-2 \ | \ \mathbf{x})$ are not guaranteed to be monotonically decreasing, which means that if we compute the discrete probabilities $p(y = 0 \ | \ \mathbf{x}), \dots, p(y = K-1 \ | \ \mathbf{x})$ these are not guaranteed to be strictly positive. To address the monotonicity issue, \citet{schapire2002modeling} proposed a heuristic solution. 

There are other ordinal techniques but which do not impose unimodal constraints. The proportional odds model (POM) and its neural network extensions (POMNN, CHNN \cite{hyperspheres}) do not suffer from the monotonicity issue due to the utilization of monotonically increasing biases in the calculation of probabilities. The stick-breaking approach by \citet{khan2012stick}, which is a reformulation of the multinomial logit (softmax), could also be used in the ordinal case as it technically imposes an ordering on classes.

\subsection{Poisson distribution}

\begin{figure}
\centering
\includegraphics[width=0.5\textwidth]{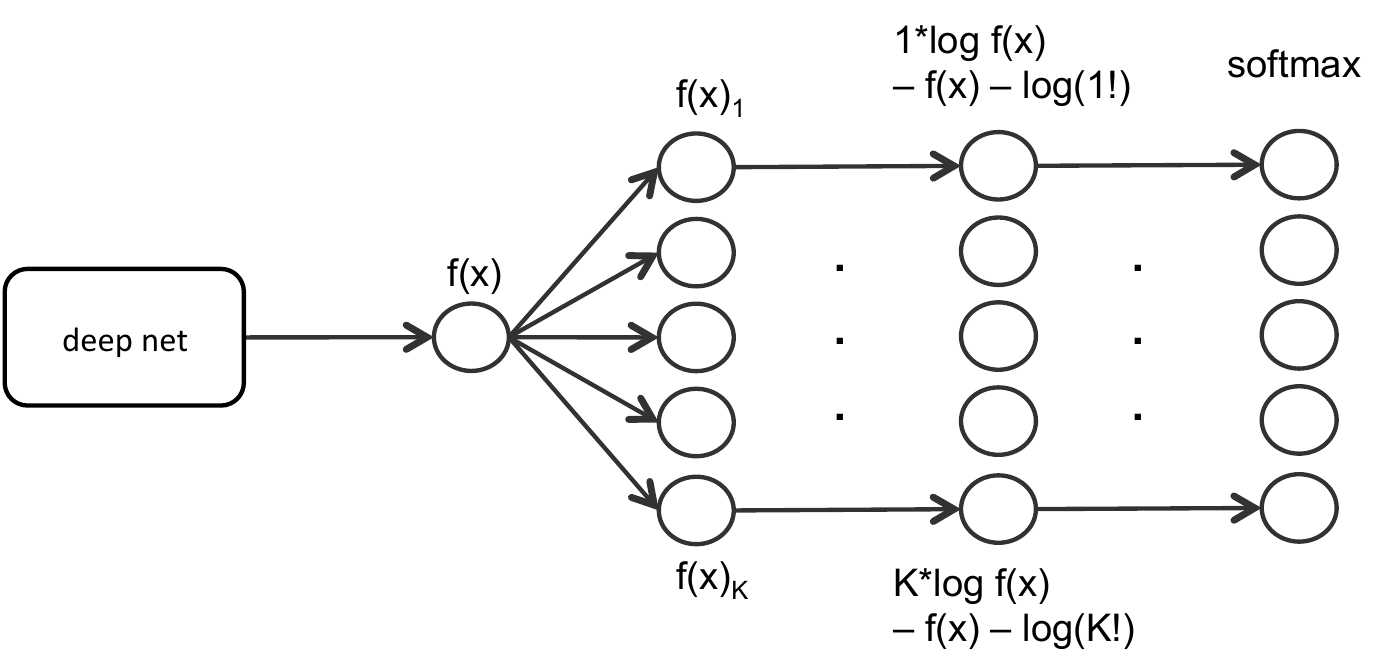}
\caption{The first layer after $f(\mathbf{x})$ is a `copy' layer, that is, $f(\mathbf{x}) = f(\mathbf{x})_{1} = \dots = f(\mathbf{x})_{K}$. The second layer applies the log Poisson PMF transform followed by the softmax layer.}
\label{fig:diagram}
\end{figure}

The Poisson distribution is commonly used to model the probability of the number of events, $k \in \mathbb{N} \cup 0$ occurring in a particular interval of time. The average frequency of these events is denoted by $\lambda \in R^{+}$. The probability mass function is defined as:
\begin{equation}
p(k; \lambda) = \frac{\lambda^{k}\exp(-\lambda)}{k!},
\end{equation}
where $0 \leq k \leq K-1$. While we are not actually using this distribution to model the occurrence of events, we can make use of its probability mass function (PMF) to enforce discrete unimodal probability distributions. For a purely technical reason, we instead deal with the log of the PMF:
\begin{equation}
\begin{split} \label{eq:log}
\log\Big[ \frac{\lambda^{k}exp(-\lambda)}{k!} \Big] &= \log(\lambda^{k}exp(-\lambda)) - \log(k!) \\
&= \log(\lambda^{k}) + \log(exp(-\lambda)) - \log(k!) \\
&= k \log(\lambda) - \lambda - \log(k!).
\end{split}
\end{equation}
If we let $f(\mathbf{x})$ denote the scalar output of our deep net (where $f(\mathbf{x}) > 0$ which can be enforced with the softplus nonlinearity), then we denote $h(\mathbf{x})_{j}$ to be:
\begin{equation} \label{eq:fx_and_bias}
j \log(f(\mathbf{x})) - f(\mathbf{x}) - \log(j!),
\end{equation}
where we have simply replaced the $\lambda$ in equation (\ref{eq:log}) with $f(\mathbf{x})$. Then, $p(y = j | \mathbf{x})$ is simply a softmax over $h(\mathbf{x})$:
\begin{equation}
p(y = j | \mathbf{x}) = \frac{\exp(-h(\mathbf{x})_{j} / \tau)}{\sum_{i=1}^{K}\exp(-h(\mathbf{x})_{i} / \tau)},
\end{equation}
which is required since the support of the Poisson is infinite. We have also introduced a hyperparameter to the softmax, $\tau$, to control the relative magnitudes of each value of $p(y = j | \mathbf{x})$ (i.e., the variance of the distribution). Note that as $\tau \rightarrow \infty$, the probability distribution becomes more uniform, and as $\tau \rightarrow 0$, the distribution becomes more `one-hot' like with respect to the class with the largest pre-softmax value.
We can illustrate this technique in terms of the layers at the end of the deep network, which is shown in Figure \ref{fig:diagram}.

We note that the term in equation (\ref{eq:fx_and_bias}) can be re-arranged and simplified to
\begin{equation} \label{eq:simpl}
\begin{split}
h(\mathbf{x})_{j} &= j \ \log(f(\mathbf{x})) - f(\mathbf{x}) - \log(j!) \\
&= -f(\mathbf{x}) + j \log(f(\mathbf{x})) - \log(j!) \\
&= -f(\mathbf{x}) + b_{j}( f(\mathbf{x}) ).
\end{split}
\end{equation}
In this form, we can see that the probability of class $j$ is determined by the scalar term $f(\mathbf{x})$ and a bias term that also depends on $f(\mathbf{x})$. Another technique that uses biases to determine class probabilities is the proportional odds model (POM), also called the ordered logit \cite{pom}, where the cumulative probability of a class depends on a learned bias:
\begin{equation} \label{eq:pom}
p(y \leq j \ | \ \mathbf{x}) = \textmd{sigm}( f(\mathbf{x}) - \mathbf{b}_{j} ),
\end{equation}
where $\mathbf{b}_{1} < \dots < \mathbf{b}_{K}$. Unlike our technique however, the bias vector $\mathbf{b}$ is not a function of $\mathbf{x}$ nor $f(\mathbf{x})$, but a fixed vector that is learned during training, which is interesting. Furthermore, probability distributions computed using this technique are not guaranteed to be unimodal.

Figure \ref{fig:k4_tau} shows the resulting probability distributions for values of $f(x) \in [0.1,4.85]$ when $\tau = 1.0$ and $\tau = 0.3$. We can see that all distributions are unimodal and that by gradually increasing $f(\mathbf{x})$ we gradually change which class has the most mass associated with itself. The $\tau$ is also an important parameter to tune as it alters the variance of the distribution. For example, in Figure \ref{fig:k4_tau_1}, we can see that if we are confident in predicting the second class, $f(\mathbf{x})$ should be $\sim 2.6$, though in this case the other classes receive almost just as much probability mass. If we set $\tau = 0.3$ however (Figure \ref{fig:k4_tau_0p3}), at $f(\mathbf{x}) = 2.6$ the second class has relatively more mass, which is to say we are even more confident that this is the correct class. An unfortunate side effect of using the Poisson distribution is that the variance is equivalent to the mean, $\lambda$. This means that in the case of a large number of classes probability mass will be widely distributed, and this can be seen in the $K = 8$ case in Figure \ref{fig:k8_tau}. While careful selection of $\tau$ can mitigate this, we also use this problem to motivate the use of the binomial distribution.

In the work of \citet{da2008unimodal}, they address the infinite support problem by using a `right-truncated' Poisson distribution. In this formulation, they simply find the normalization constant such that the probabilities sum to one. This is almost equivalent to what we do, since we use a softmax, although the softmax exponentiates its inputs and we also introduce the temperature parameter $\tau$ to control for the variance of the distribution.


\subsection{Binomial distribution}

The binomial distribution is used to model the probability of a given number of `successes' out of a given number of trials and some success probability. The probability mass function for this distribution -- for $k$ successes (where $0 \leq k \leq K-1$), given $K-1$ trials and success probability $p$ -- is:
\begin{equation}
p(k; K-1, p) = \binom{K-1}{k} p^{k}(1 - p)^{K-1-k}
\end{equation}
In the context of applying this to a neural network, $k$ denotes the class we wish to predict, $K-1$ denotes the number of classes (minus one since we index from zero), and $p = f(\mathbf{x}) \in [0,1]$ is the output of the network that we wish to estimate. While no normalisation is theoretically needed since the binomial distribution's support is finite, we still had to take the log of the PMF and normalise with a softmax to address numeric stability issues. This means the resulting network is equivalent to that shown in Figure \ref{fig:diagram}, but with the log binomial PMF instead of Poisson. Just like with the Poisson formulation, we can introduce the temperature term $\tau$ into the resulting softmax to control for the variance of the resulting distribution.

\begin{figure*}
    \centering
    \subfigure[$K = 4$]{\label{fig:binom_k4}\includegraphics[width=0.45\textwidth]{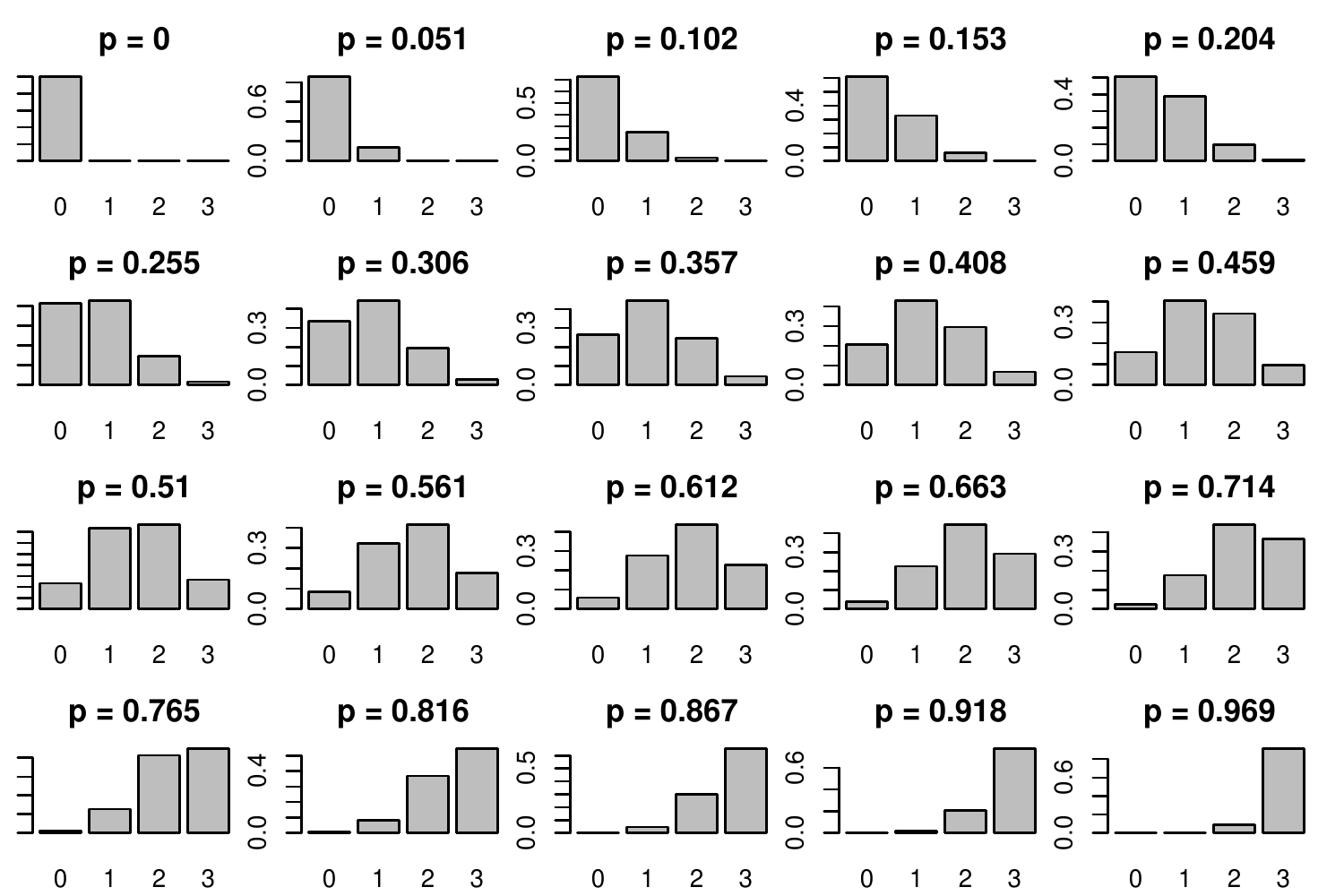}}  \subfigure[$K = 8$]{\label{fig:binom_k8}\includegraphics[width=0.45\textwidth]{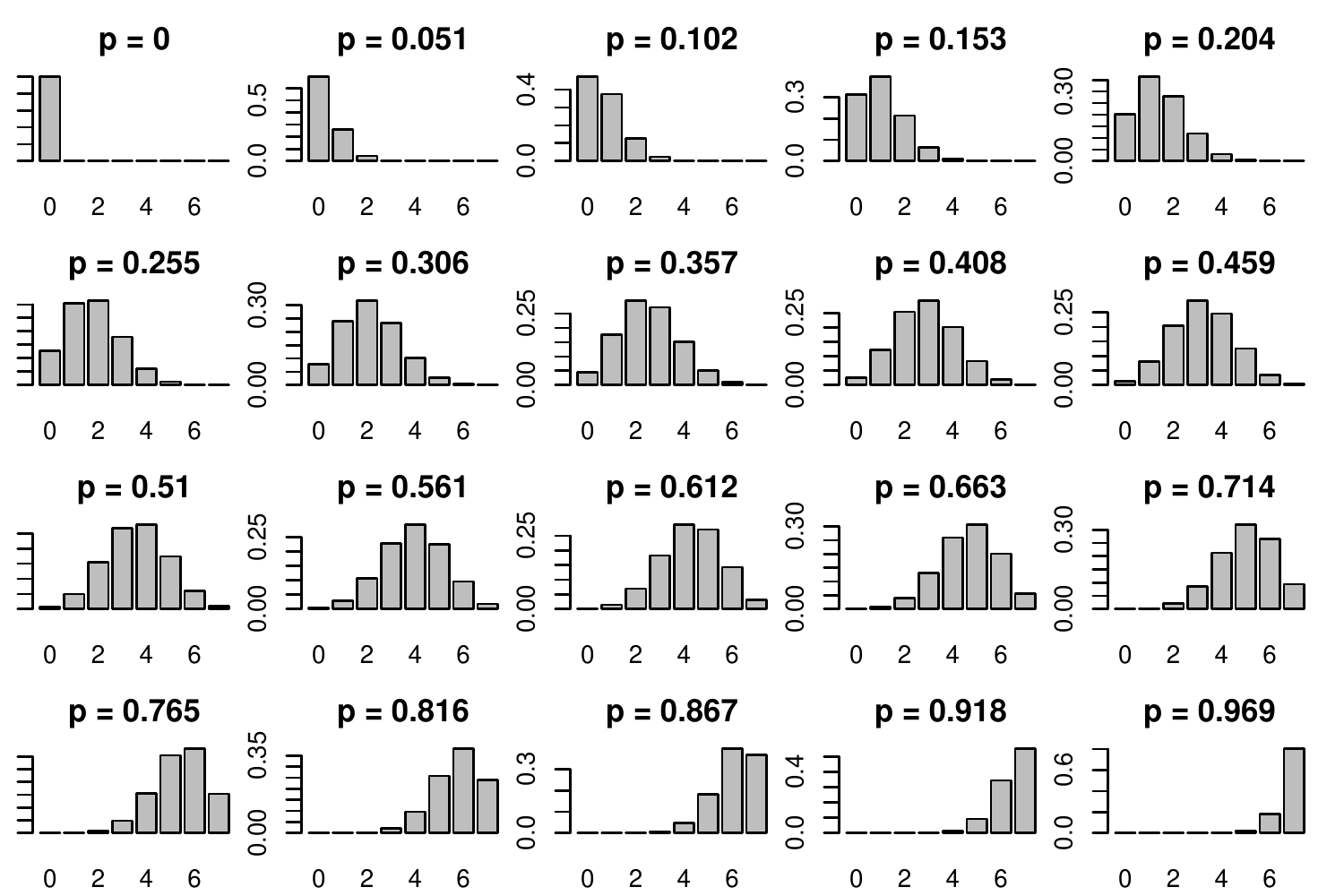}} 
    \caption{Illustration of the probability distributions that are obtained from varying $p \in [0,1]$ for the binomial classes when $K = 4$ (left) and $K = 8$ (right).}
    \label{fig:binomd}
\end{figure*}

Figure \ref{fig:binomd} shows the resulting distributions achieved by varying $p$ for when $K = 4$ and $K = 8$.

\section{Methods and Results}

In this section we go into details of our experiments, including the datasets used and the precise architectures.

\subsection{Data}

We make use of two ordinal datasets appropriate for deep neural networks:
\begin{itemize}
\item Diabetic retinopathy\footnote{https://www.kaggle.com/c/diabetic-retinopathy-detection/}. This is a dataset consisting of extremely high-resolution fundus image data. The training set consists of 17,563 pairs of images (where a pair consists of a left and right eye image corresponding to a patient). In this dataset, we try and predict from five levels of diabetic retinopathy: no DR (25,810 images), mild DR (2,443 images), moderate DR (5,292 images), severe DR (873 images), or proliferative DR (708 images). A validation set is set aside, consisting of 10\% of the patients in the training set. The images are pre-processed using the technique proposed by competition winner \citet{report} and subsequently resized to 256px width and height.
\item The Adience face dataset\footnote{http://www.openu.ac.il/home/hassner/Adience/data.html} \cite{eidinger2014age}. This dataset consists of 26,580 faces belonging to 2,284 subjects. We use the form of the dataset where faces have been pre-cropped and aligned. We further pre-process the dataset so that the images are 256px in width and height. The training set consists of merging the first four cross-validation folds together (the last cross-validation fold is the test set), which comprises a total of 15,554 images. From this, 10\% of the images are held out as part of a validation set.
\end{itemize}

\subsection{Network}

We make use of a modest ResNet \cite{resnets} architecture to conduct our experiments. Table \ref{tab:architecture} describes the exact architecture. We use the ReLU nonlinearity and HeNormal initialization throughout the network.

\begin{table}[H]
\centering
\begin{tabular}{|l|l|}
\hline
\textbf{Layer}                        & \textbf{Output size}  \\ \hline
Input (3x224x224)                     & 3 x 224 x 224  \\ \hline
Conv (7x7@32s2)                       & 32 x 112 x 112  \\ \hline
MaxPool (3x3s2)                       & 32 x 55 x 55 \\ \hline
2 x ResBlock (3x3@64s1)               & 32 x 55 x 55  \\ \hline
1 x ResBlock (3x3@128s2)               & 64 x 28 x 28 \\ \hline
2 x ResBlock (3x3@128s1)               & 64 x 28 x 28  \\ \hline
1 x ResBlock (3x3@256s2)              & 128 x 14 x 14 \\ \hline
2 x ResBlock (3x3@256s1)              & 128 x 14 x 14 \\ \hline
1 x ResBlock (3x3@512s2)              & 256 x 7 x 7 \\ \hline
2 x ResBlock (3x3@512s1)              & 256 x 7 x 7 \\ \hline
AveragePool (7x7s7)                   & 256 x 1 x 1     \\ \hline
\end{tabular}
\caption{Description of the ResNet architecture used in the experiments. For convolution, WxH@FsS = filter size of dimension W x H, with F feature maps, and a stride of S. For average pooling, WxHsS = a pool size of dimension W x H with a stride of S. This architecture comprises a total of 4,307,840 learnable parameters.}
\label{tab:architecture}
\end{table}


We conduct the following experiments for both DR and Adience datasets:
\begin{itemize}
\item (Baseline) cross-entropy loss. This simply corresponds to a softmax layer for $K$ classes at the end of the average pooling layer in Table \ref{tab:architecture}. For Adience and DR, this corresponds to a network with 4,309,896 and 4,309,125 learnable parameters, respectively.
\item (Baseline) squared-error loss. Rather than regress $f(\mathbf{x})$ against $y$, we regress with $(K-1)sigm(f(\mathbf{x}))$, since we have observed better results with this formulation in the past. For Adience and DR, this corresponds t 4,309,905 and 4,309,131 learnable parameters, respectively.
\item Cross-entropy loss using the Poisson and binomial extensions at the end of the architecture (see Figure \ref{fig:diagram}). For Adience and DR, this corresponds to 4,308,097 learnable parameters for both. Although \citet{da2008unimodal} mention that cross-entropy or squared error can be used, their equations assume a squared error between the (one-hot encoded) ground truth and $p(y|\mathbf{x})$, whereas we use cross-entropy.
\item EMD loss (equation \ref{eq:emd2}) where $\ell = 2$ (i.e. Euclidean norm) and the entire term is squared (to get rid of the square root induced by the norm) using Poisson and binomial extensions at the end of architecture. Again, this corresponds to 4,308,097 learnable parameters for both networks.
\end{itemize}

Amongst these experiments, we use $\tau = 1$ and also learn $\tau$ as a bias. When we learn $\tau$, we instead learn $\textmd{sigm}(\tau)$ since we found this made training more stable. Note that we can also go one step further and learn $\tau$ as a function of $\mathbf{x}$, though experiments did not show any significant gain over simply learning it as a bias. However, one advantage of this technique is that the network can quantify its uncertainty on a per-example basis. It is also worth noting that the Poisson and binomial formulations are slightly underparameterised compared to their baselines, but experiments we ran that addressed this (by matching model capacity) did not yield significantly different results.

It is also important to note that in the case of ordinal prediction, there are two ways to compute the final prediction: simply taking the argmax of $p(y|\mathbf{x})$ (which is what is simply done in discrete classification), or taking a `smoothed' prediction which is simply the expectation of the integer labels w.r.t. the probability distribution, i.e., $\mathbb{E}[0, \dots, K-1]_{p(y|x)}$. For the latter, we call this the `expectation trick'. A benefit of the latter is that it computes a prediction that considers the probability mass of all classes. One benefit of the former however is that we can use it to easily rank our predictions, which can be important if we are interested in computing top-$k$ accuracy (rather than top-1).

We also introduce an ordinal evaluation metric -- the quadratic weighted kappa (QWK) \cite{cohen1968weighted} -- which has seen recent use on ordinal competitions on Kaggle. Intuitively, this is a number between [-1,1], where a kappa $\kappa = 0$ denotes the model does no better than random chance, $\kappa < 0$ denotes worst than random chance, and $\kappa > 0$ better than random chance (with $\kappa = 1$ being the best score). The `quadratic' part of the metric imposes a quadratic penalty on misclassifications, making it an appropriate metric to use for ordinal problems.\footnote{The quadratic penalty is arbitrary but somewhat appropriate for ordinal problems. One can plug in any cost matrix into the kappa calculation.}

All experiments utilise an $\ell_{2}$ norm of $10^{-4}$, ADAM optimiser \cite{adam} with initial learning rate $10^{-3}$, and batch size 128. A `manual' learning rate schedule is employed where we manually divide the learning rate by 10 when either the validation loss or valid set QWK plateaus (whichever plateaus last) down to a minimum of $10^{-4}$ for Adience and $10^{-5}$ for DR.\footnote{We also re-ran experiments using an automatic heuristic to change the learning rate, and similar experimental results were obtained.}

\begin{figure*}
    \centering
    \subfigure[Learning curves for Adience dataset, for $\tau = 1.0$. For both accuracy and QWK, both the argmax and expectation way of computing a prediction are employed.]{\label{fig:ad_curves_a}\includegraphics[width=0.95\textwidth]{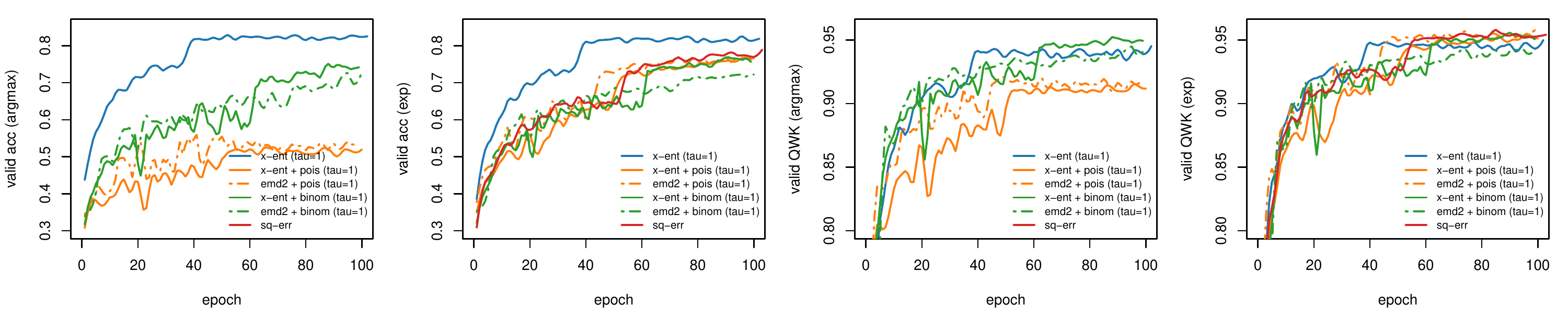}}
    ~
    \subfigure[Learning curves for Adience dataset, for when $\tau$ is learned as a bias. For both accuracy and QWK, both the argmax and expectation way of computing a prediction are employed.]{\label{fig:ad_curves_c}\includegraphics[width=0.95\textwidth]{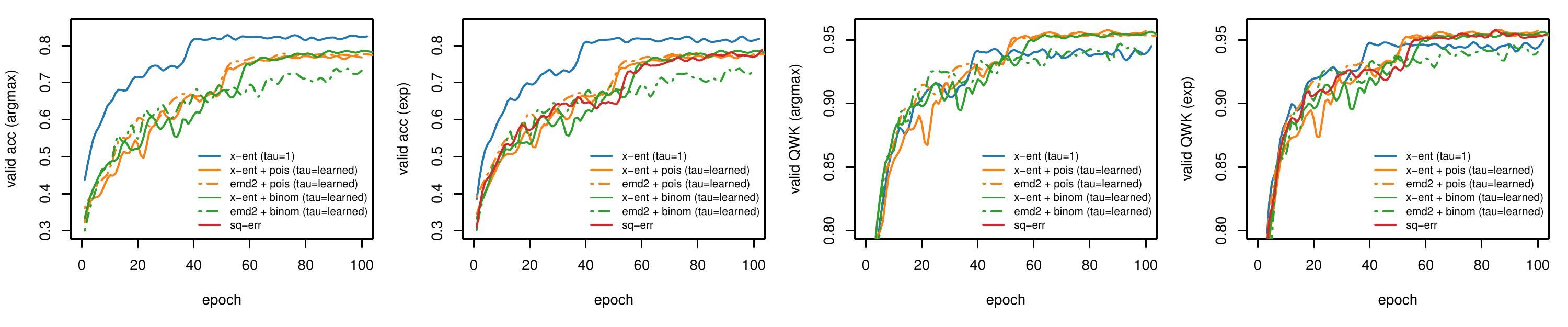}}
    \caption{Experiments for the Adience dataset. For $\tau = 1$ and $\tau = \text{learned}$, we compare typical cross-entropy loss (blue), cross-entropy/EMD with Poisson formulation (orange solid / dashed, respectively), cross-entropy/EMD with binomial formulation (green solid / dashed, respectively), and regression (red). Learning curves have been smoothed with a LOESS regression for presentation purposes.}
    \label{fig:ad_curves}
\end{figure*}


\begin{figure*}
    \centering
    \subfigure[Learning curves for diabetic retinopathy dataset, for $\tau = 1.0$. For both accuracy and QWK, both the argmax and expectation way of computing a prediction are employed.]{\label{fig:dr_curves_a}\includegraphics[width=0.95\textwidth]{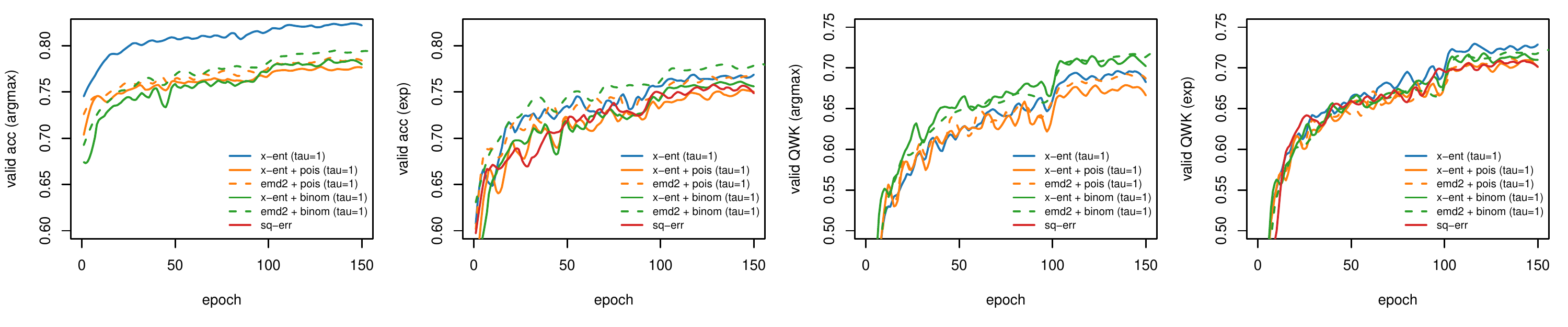}}    
    ~
    \subfigure[Learning curves for diabetic retinopathy dataset, where $\tau$ is made a learnable bias. For both accuracy and QWK, both the argmax and expectation way of computing a prediction are employed.]{\label{fig:dr_curves_c}\includegraphics[width=0.95\textwidth]{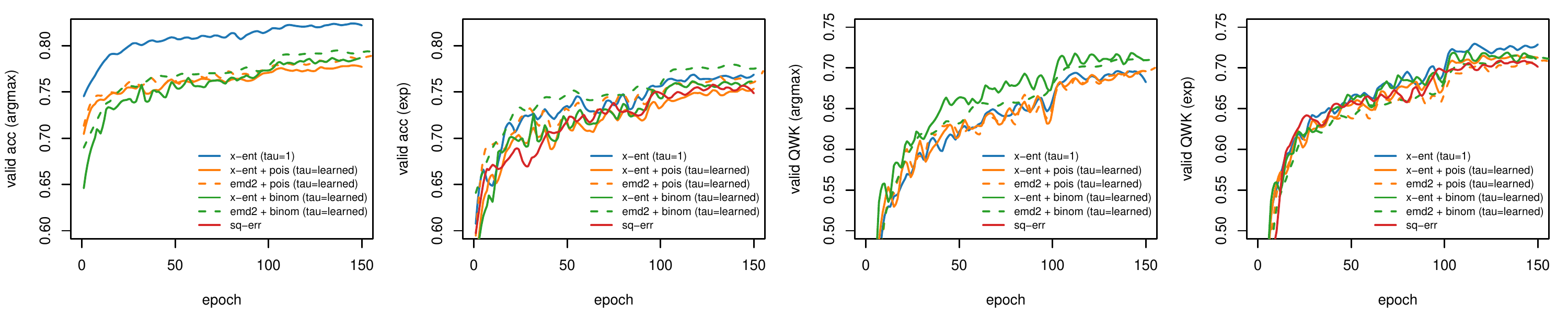}}    
    \caption{Experiments for the diabetic retinopathy (DR) dataset. For $\tau = 1$ and $\tau = \text{learned}$, we compare typical cross-entropy loss (blue), cross-entropy/EMD with Poisson formulation (orange solid / dashed, respectively), cross-entropy/EMD with binomial formulation (green solid / dashed, respectively), and regression (red). Learning curves have been smoothed with a LOESS regression for presentation purposes.}
    \label{fig:dr_curves}
\end{figure*}

\subsection{Experiments}

\begin{figure*}
    \centering
    \subfigure[$K = 4, \tau = 1.0$]{\label{fig:k4_tau_1}\includegraphics[width=0.45\textwidth]{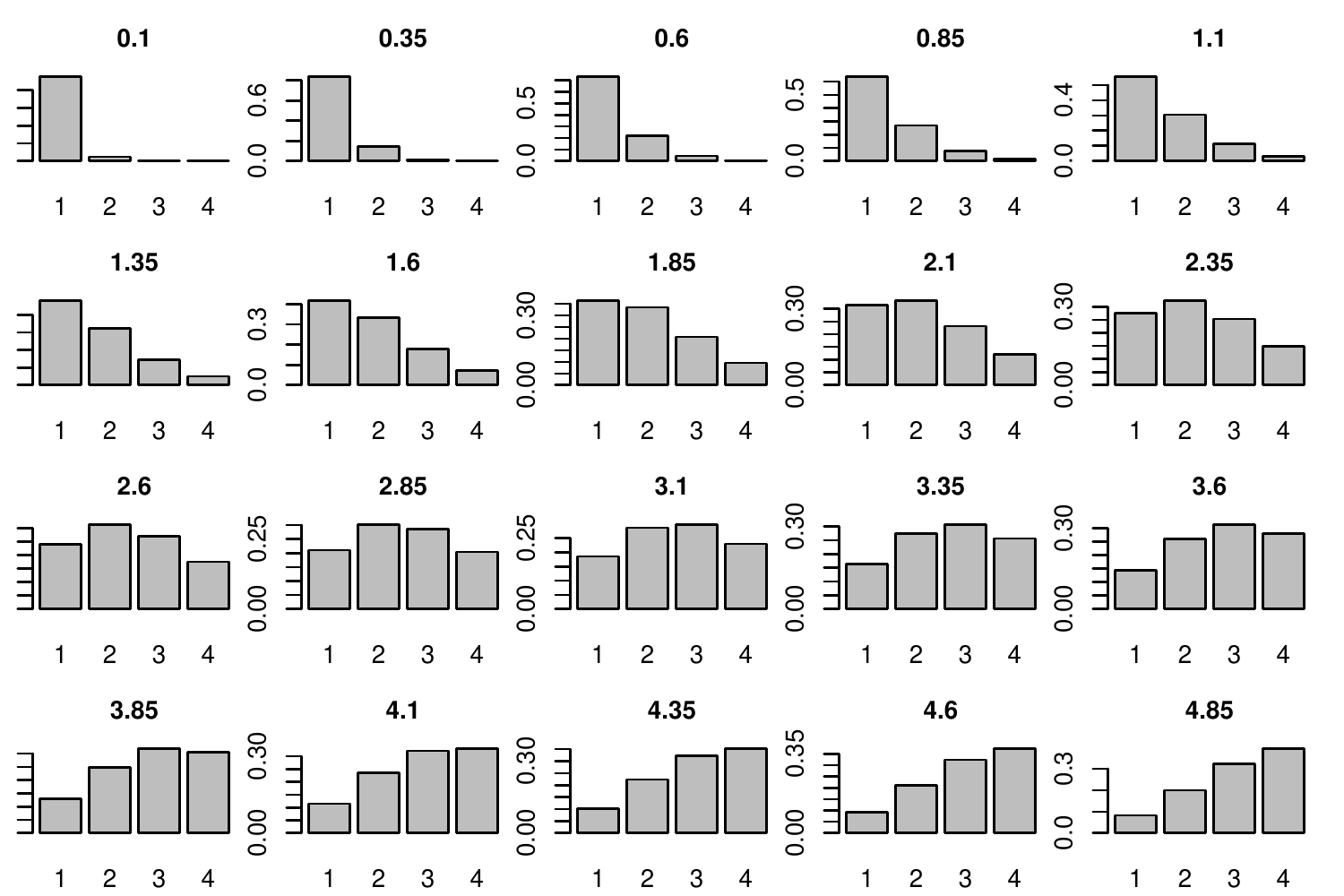}}  \subfigure[$K = 4, \tau = 0.3$]{\label{fig:k4_tau_0p3}\includegraphics[width=0.45\textwidth]{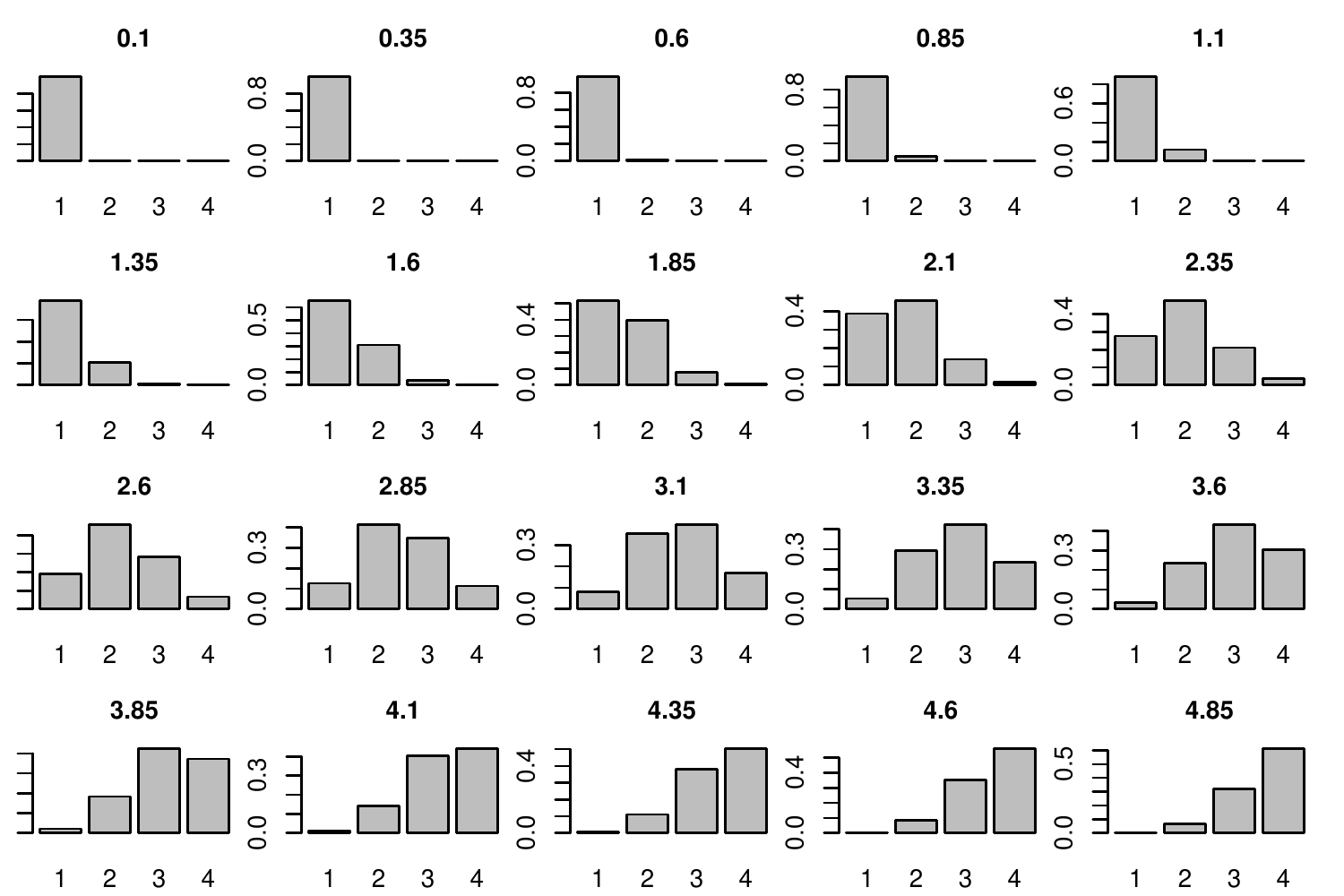}} 
    \caption{Illustration of the probability distributions that are obtained from varying $f(\mathbf{x}) \in [0.1, 4.85]$ for when there are four classes ($K = 4$) and when $\tau = 1.0$ (left) and $\tau = 0.3$ (right). We can see that lowering $\tau$ results in a lower variance distribution. Depending on the number of classes, it may be necessary to tune $\tau$ to ensure the right amount of probability mass hits the correct class.}
    \label{fig:k4_tau}
\end{figure*}

\begin{figure*}
    \centering
    \subfigure[$K = 8, \tau = 1.0$]{\label{fig:k8_tau_1}\includegraphics[width=0.45\textwidth]{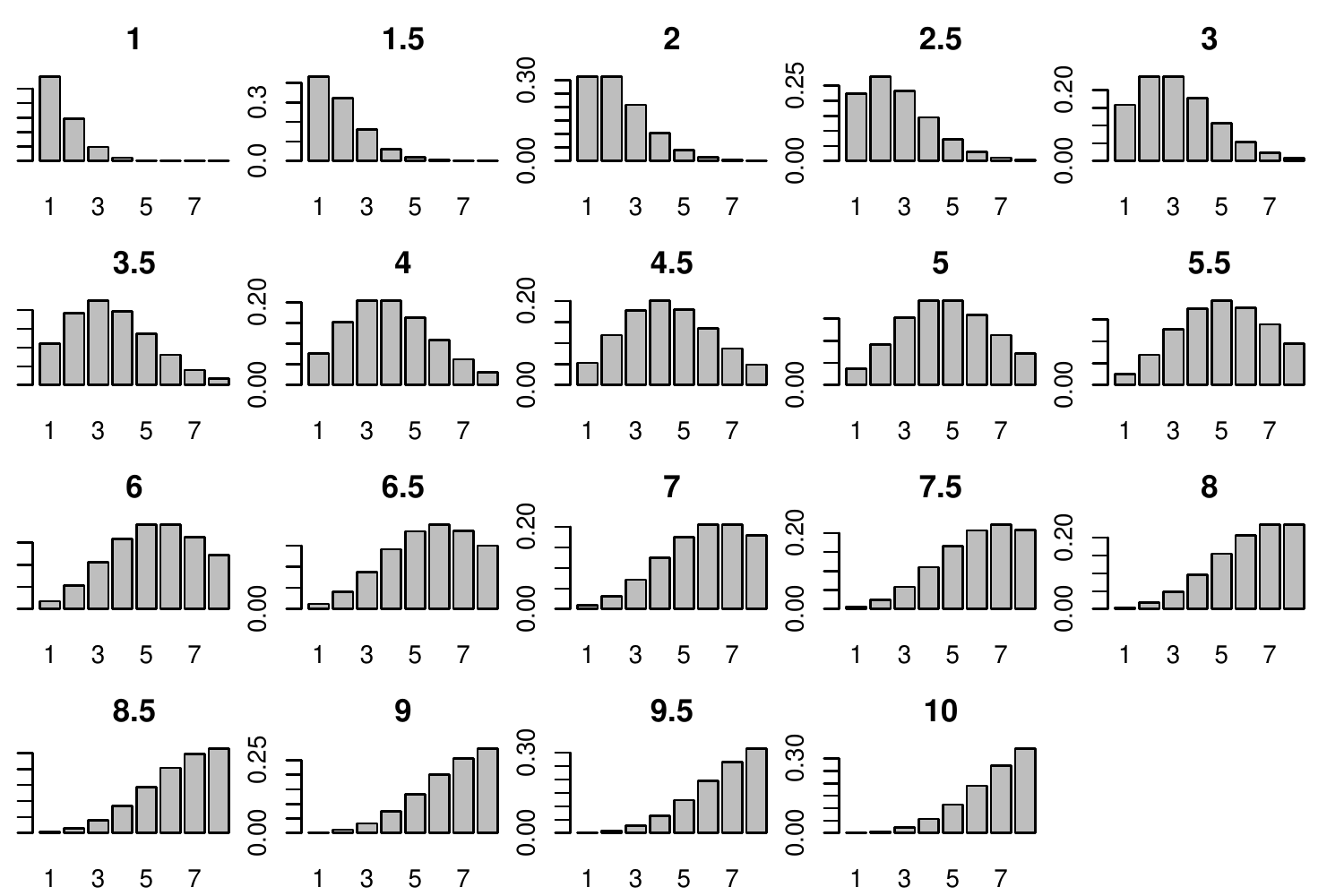}}  \subfigure[$K = 8, \tau = 0.3$]{\label{fig:k8_tau_0p3}\includegraphics[width=0.45\textwidth]{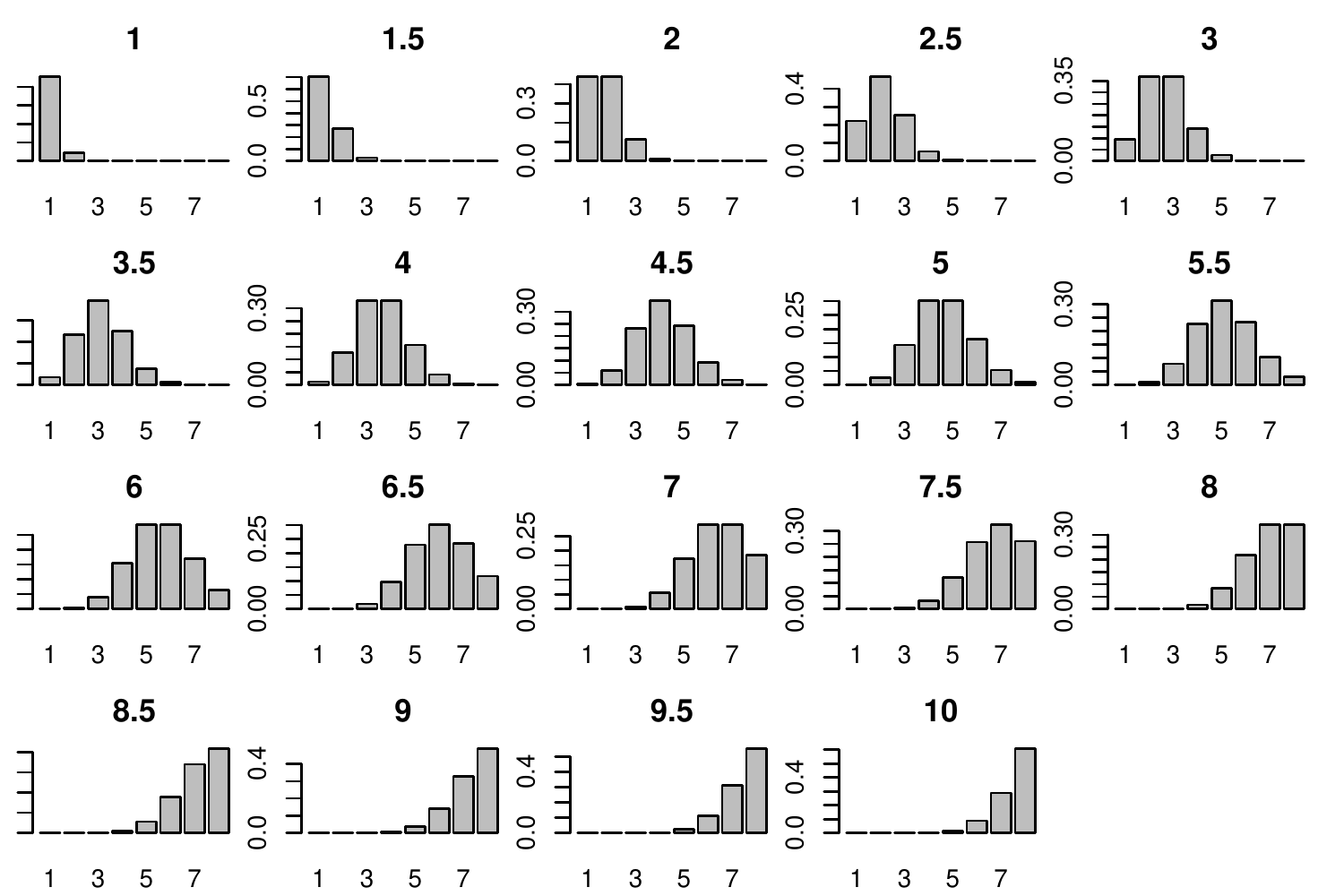}} 
    \caption{Illustration of the probability distributions that are obtained from varying $f(\mathbf{x}) \in [1,10]$ for when there are eight classes ($K = 8$) and when $\tau = 1.0$ (left) and $\tau = 0.3$ (right). Because we have a greater number of classes, $f(\mathbf{x})$ must take on a greater range of values (unlike Figure \ref{fig:k4_tau}) before most of the probability mass moves to the last class.}
    \label{fig:k8_tau}
\end{figure*}

\begin{figure*}
    \centering
    \subfigure[Faces from Adience valid set]{\label{fig:faces}\includegraphics[width=0.32\textwidth]{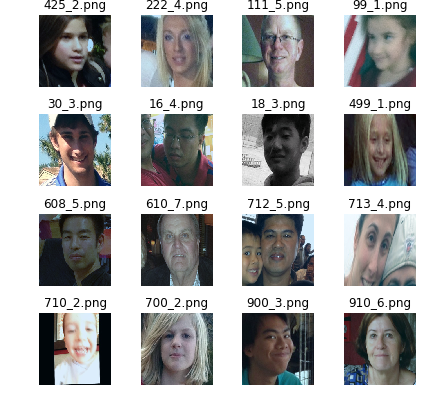}} \subfigure[Cross-entropy (baseline)]{\label{fig:adience_xent_dists}\includegraphics[width=0.32\textwidth]{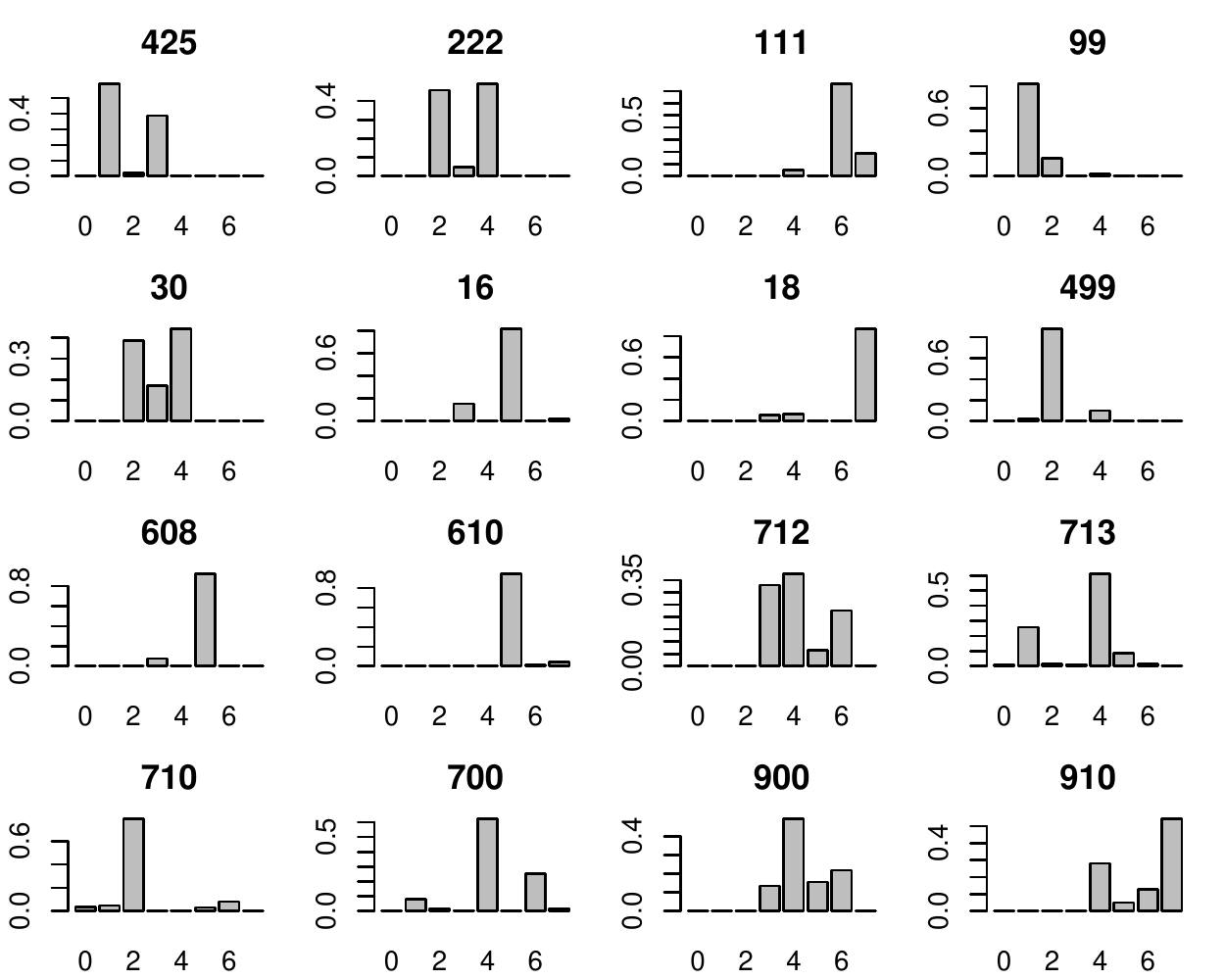}}  \subfigure[Cross-entropy + Poisson ($\tau = 1.0$)]{\label{fig:adience_pois_dists}\includegraphics[width=0.32\textwidth]{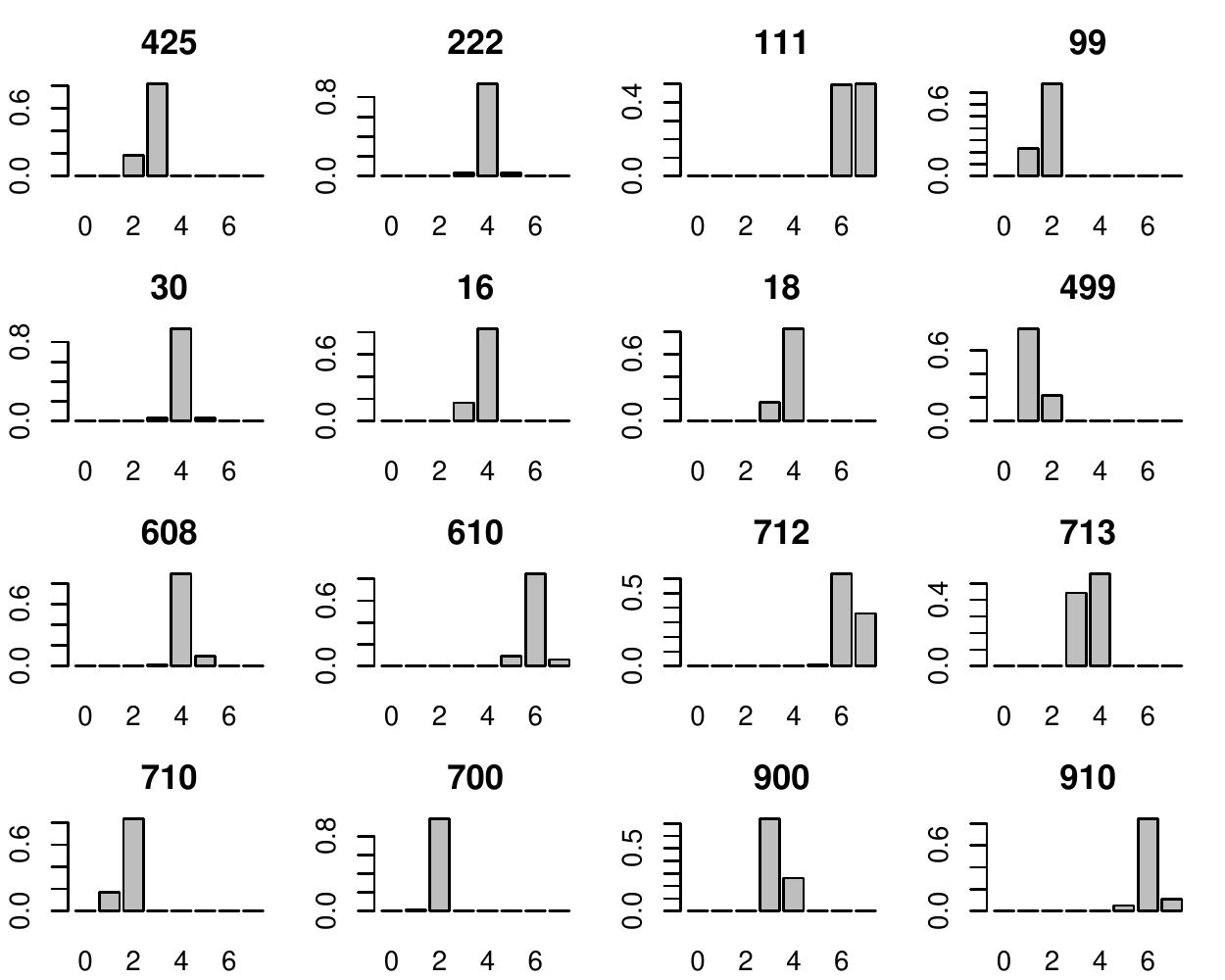}} 
    \caption{Probability distributions over selected examples in the validation set for Adience (those selected have non-unimodal probability distributions for the cross-entropy baseline). Left: from cross-entropy + Poisson model ($\tau$ learned), right: cross-entropy (baseline) model}
    \label{fig:adience_dists}
\end{figure*}

Figure \ref{fig:ad_curves} shows the experiments run for the Adience dataset, for when $\tau = 1.0$ (Figure \ref{fig:ad_curves_a}) and when $\tau$ is learned (Figure \ref{fig:ad_curves_c}). We can see that for our methods, careful selection of $\tau$ is necessary for the accuracy on the validation set to be on par with that of the cross-entropy baseline. For $\tau = 1.0$, accuracy is poor, but even less so when $\tau$ is learned. To some extent, using the smoothed prediction with the expectation trick alleviates this gap. However, because the dataset is ordinal, accuracy can be very misleading, so we should also consider the QWK. For both argmax and expectation, our methods either outperform or are quite competitive with the baselines, with the exception of the QWK argmax plot for when $\tau = 1$, where only our binomial formulations were competitive with the cross-entropy baseline. Overall, considering all plots in Figure \ref{fig:ad_curves} it appears the binomial formulation produces better results than Poisson. There also appears to be some benefit gained from using the EMD loss for Poisson, but not for binomial.

Figure \ref{fig:dr_curves} show the experiments run for diabetic retinopathy. We note that unlike Adience, the validation accuracy does not appear to be so affected across all specifications of $\tau$. One potential reason for this is due to Adience having a larger number of classes compared to DR. As we mentioned earlier, the Poisson distribution is somewhat awkward as its variance is equivalent to its mean. Since most of the probability mass sits at the mean, if the mean of the distribution is very high (which is the case for datasets with a large $K$ such as Adience), then the large variance can negatively impact the distribution by taking probability mass away from the correct class. We can see this effect by comparing the distributions in Figure \ref{fig:k4_tau} ($k = 4$) and Figure \ref{fig:k8_tau} ($k = 8$). As with the Adience dataset, the use of the expectation trick brings the accuracy of our methods to be almost on-par with the baselines. In terms of QWK, only our binomial formulations appear to be competitive, but only in the argmax case do one of our methods (the binomial formulation) beat the cross-entropy baseline. At least for accuracy, there appears to be some gain in using the EMD loss for the binomial formulation. Because DR is a much larger dataset compared to Adience, it is possible that the deep net is able to learn reasonable and `unimodal-like' probability distributions without it being enforced in the model architecture.

\begin{figure}[H]
\centering
\includegraphics[width=0.45\textwidth]{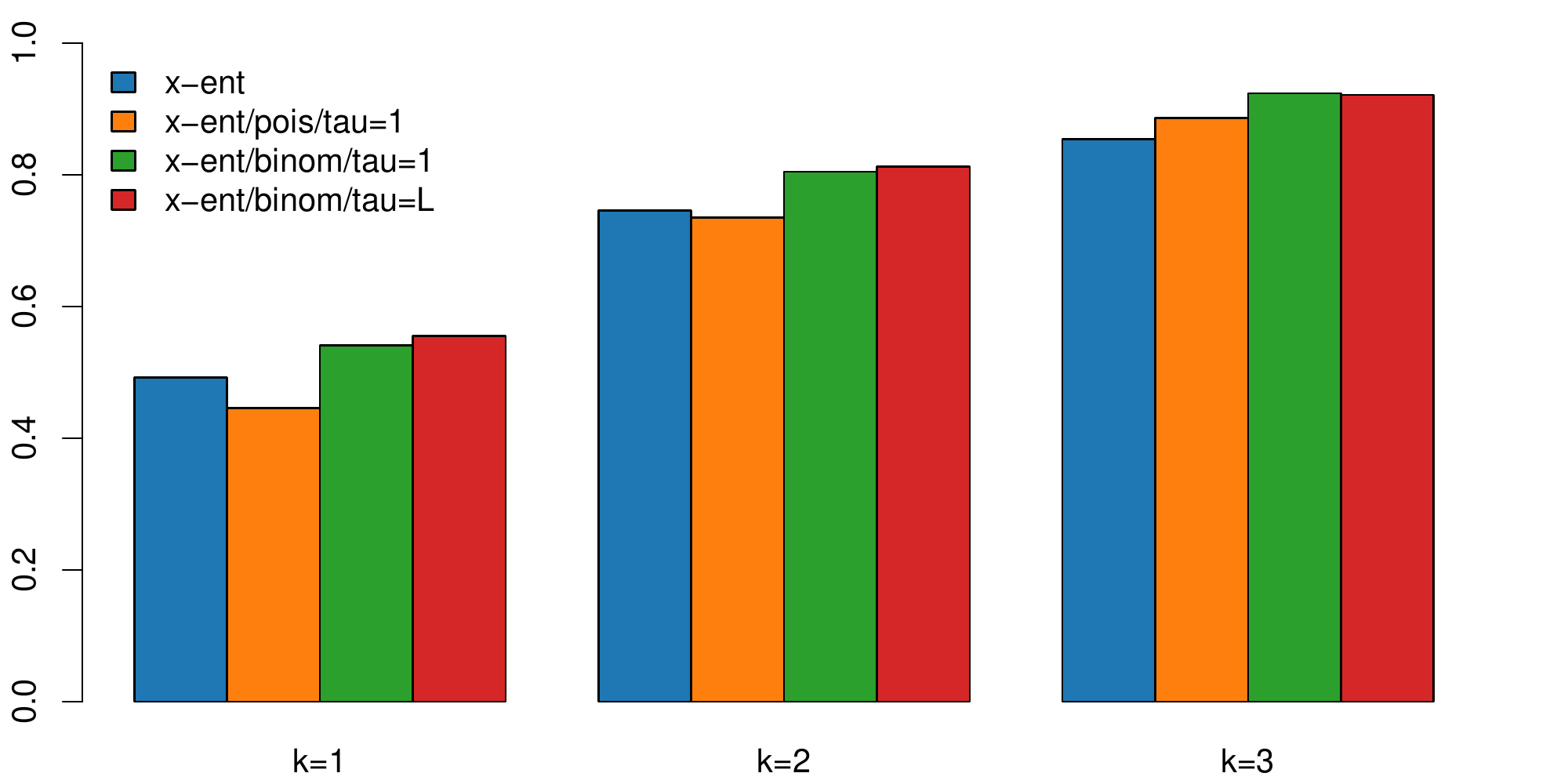}
\caption{Top-$k$ accuracies computed on the Adience test set, where $k \in \{1, 2, 3\}$. }
\label{fig:adience_test}
\end{figure}

Overall, across both datasets the QWK for our methods are generally at least competitive with the baselines, especially if we learn $\tau$ to control for the variance. In the empirical results of \citet{da2008unimodal}, they found that the binomial formulation performed better than the Poisson, and when we consider all of our results in Figure \ref{fig:ad_curves} and \ref{fig:dr_curves} we come to the same conclusion. They justify this result by defining the `flexibility' of a discrete probability distribution and show that the binomial distribution is more `flexible' than Poisson. From our results, we believe that these unimodal methods act as a form of regularization which can be useful in regimes where one is interested in top-$k$ accuracy. For example, in the case of top-$k$ accuracy, we want to know if the ground truth was in the top $k$ predictions, and we may be interested in such metrics if it is difficult to achieve good top-1 accuracy. Assume that our probability distribution $p(y|\mathbf{x})$ has most of its mass on the wrong class, but the correct class is on either side of it. Under a unimodal constraint, it is guaranteed that the two classes on either side of the majority class will receive the next greatest amount of probability mass, and this can result in a correct prediction if we consider top-2 or top-3 accuracy. To illustrate this, we compute the top-$k$ accuracy on the test set of the Adience dataset, shown in Figure \ref{fig:adience_test}. We can see that even with the worst-performing model -- the Poisson formulation with $\tau = 1$ (orange) -- produces a better top-3 accuracy than the cross-entropy baseline (blue).

\section{Conclusion}

In conclusion, we present a simple technique to enforce unimodal ordinal probabilistic predictions through the use of the binomial and Poisson distributions. This is an important property to consider in ordinal classification because of the inherent ordering between classes. We evaluate our technique on two ordinal image datasets and obtain results competitive or superior to the cross-entropy baseline for both the quadratic weighted kappa (QWK) metric and top-$k$ accuracy for both cross-entropy and EMD losses, especially under the binomial distribution. Lastly, the unimodal constraint can makes the probability distributions behave more sensibly in certain settings. However, there may be ordinal problems where a multimodal distribution may be more appropriate. We leave an exploration of this issue for future work. Code will be made available here.\footnote{https://github.com/christopher-beckham/deep-unimodal-ordinal}

\section{Acknowledgements}

We thank Samsung for funding this research. We would like to thank the contributors of Theano \cite{theano} and Lasagne \cite{lasagne} (which this project was developed in predominantly), as well as Keras \cite{keras} for extra useful code. We thank the ICML reviewers for useful feedback, as well as Eibe Frank.

\pagebreak

\bibliography{main}
\bibliographystyle{icml2017}

\end{document}